\documentclass{article}

 \usepackage[preprint]{neurips_2025}

\usepackage[utf8]{inputenc} 
\usepackage[T1]{fontenc}    
\usepackage{hyperref}       
\usepackage{url}            
\usepackage{booktabs}       
\usepackage{amsfonts}       
\usepackage{nicefrac}       
\usepackage{microtype}      
\usepackage{xcolor}         
\usepackage{graphicx}
\usepackage{subcaption}
\usepackage{amsmath}
\usepackage{multirow}  
\usepackage{array} 
\usepackage{booktabs}
\renewcommand{\arraystretch}{1.4}  

\title{2DNMRGym: An Annotated Experimental Dataset for Atom-Level Molecular Representation Learning in 2D NMR via Surrogate Supervision}

\author{%
  Yunrui Li$^{\dag}$ \\Department of Computer Science\\ Brandeis University\\
  \And Hao Xu$^{\dag}$ \\Department of Medicine\\ Harvard Medical School\\
  \And Pengyu Hong\thanks{Corresponding author: hongpeng@brandeis.edu} \\
  Department of Computer Science\\ Brandeis University\\
  $^{\dag}$ Equal contribution
}

\begin{document}

\maketitle

\begin{abstract}

Two-dimensional (2D) Nuclear Magnetic Resonance (NMR) spectroscopy, particularly Heteronuclear Single Quantum Coherence (HSQC) spectroscopy, plays a critical role in elucidating molecular structures, interactions, and electronic properties. However, accurately interpreting 2D NMR data remains labor-intensive and error-prone, requiring highly trained domain experts, especially for complex molecules. Machine Learning (ML) holds significant potential in 2D NMR analysis by learning molecular representations and  recognizing complex patterns from data. However, progress has been limited by the lack of large-scale and high-quality annotated datasets. In this work, we introduce \textbf{2DNMRGym}, the first annotated experimental dataset designed for ML-based molecular representation learning in 2D NMR. It includes over 22,000 HSQC spectra, along with the corresponding molecular graphs and SMILES strings. Uniquely, 2DNMRGym adopts a surrogate supervision setup: models are trained using algorithm-generated annotations derived from a previously validated method and evaluated on a held-out set of human-annotated gold-standard labels. This enables rigorous assessment of a model’s ability to generalize from imperfect supervision to expert-level interpretation. We provide benchmark results using a series of 2D and 3D GNN and GNN transformer models, establishing a strong foundation for future work. 2DNMRGym supports scalable model training and introduces a chemically meaningful benchmark for evaluating atom-level molecular representations in NMR-guided structural tasks. Our data and code is open-source and available at: \href{https://github.com/siriusxiao62/2DNMRGym}{https://github.com/siriusxiao62/2DNMRGym}.

\end{abstract}

\section{Introduction}
\label{intro}
\subsection{Overview}

Nuclear Magnetic Resonance (NMR) spectroscopy is a powerful technique that uses the magnetic properties of atomic nuclei to provide detailed insights into the structure and dynamics of chemical compounds \citep{gunther1994nmr, claridge2016high, yu2021recent}.
It can determine the types, quantities, and spatial arrangements of atoms within molecules and their surrounding chemical environments, from small molecules to material polymers and complex bio-macromolecules. In NMR spectrum analysis, chemists utilize prediction tools to generate chemical shifts from molecular structures, comparing them with experimental values to verify structural assignments. This comparison aids in assessing the accuracy of proposed molecular structures and provides insights into the electronic and spatial environments of atoms within the molecule.


Among NMR techniques, Heteronuclear Single Quantum Coherence (HSQC) spectroscopy \citep{bodenhausen1980natural} stands out as a powerful two-dimensional (2D) Nuclear Magnetic Resonance (NMR) method that has become indispensable for the structural elucidation of complex molecules, especially when traditional one-dimensional (1D) NMR techniques are insufficient \citep{bross2005strategies, li2020practical}.
By correlating the chemical shifts of proton nuclei with those of heteronuclei, typically \( ^{13}C \) or \( ^{15}N \), via scalar coupling interactions, HSQC enables the precise mapping of interatomic linkages within molecular frameworks. This method is particularly valuable for identifying connectivity patterns between protons and adjacent heteronuclei, thereby providing critical insights into chemical bonding, stereochemistry, and three-dimensional molecular conformation.

Despite recent advancements in the prediction of 1D NMR spectra \citep{kwon2020neural, yang2021predicting, han2022scalable, chen2024gt} and peak assignment \citep{xu2023enhancing}, the application of machine learning techniques to 2D NMR, such as HSQC spectra prediction, remains constrained by the scarcity of annotated datasets for training.
To the best of our knowledge, no large-scale annotated dataset of experimental HSQC spectra is currently available for training machine learning models. This is primarily due to the significant bottleneck in acquiring, processing, and annotating 2D NMR data. Acquiring HSQC spectra is time-consuming, requires highly sensitive instrumentation, and depends on the availability of pure samples at an appropriate concentration, making the process highly labor-intensive. Typically, a research group can only produce 10-20 high-quality spectra per week. Furthermore, the complexity of molecular structures leads to spectral overlap and signal degeneracy, complicating peak resolution. The presence of multiple chiral centers in molecules can further complicate annotations. Experimental conditions also play a critical role in determining the quality of HSQC spectra. Consequently, the requirement for expensive instruments, labor-intensive sample preparation, and specialized expertise in organic chemistry severely limit the availability of large, annotated datasets.

\begin{figure}[ht!]
    \centering
    \includegraphics[width=\textwidth]{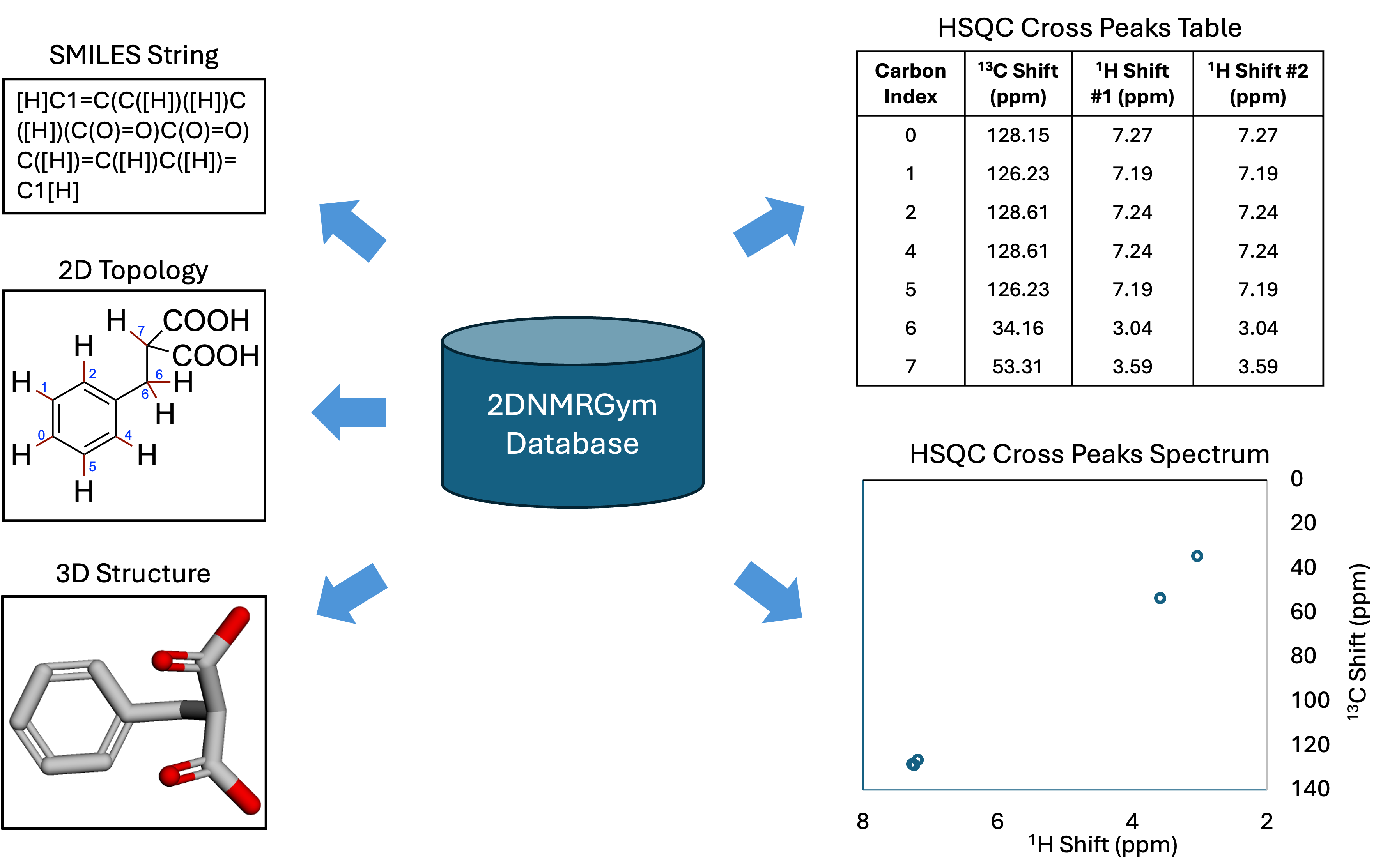}
    \caption{The 2DNMRGym dataset comprises multi-modal components, including the SMILES representation of each molecule and its conversion to a molecular graph. This graph includes both 2D topological structures and Cartesian coordinates for 3D spatial information. The ground truth spectrum is represented as cross peak tables, where the ``Carbon Index'' maps to the corresponding carbons in the molecular topology graph.}
    \label{fig:data}
\end{figure}

To fill this gap, we introduce the 2DNMRGym dataset (illustrated in Figure \ref{fig:data}), including 22,348 experimental HSQC spectra. Among these, 21,869 HSQC spectra with 33,8370 cross peaks were annotated using a recently published algorithm~\citep{li2025transpeaknet} and 479 spectra with 7,310 peaks were manually annotated and cross-validated by three domain experts. Each spectrum includes cross peaks annotated with their corresponding molecular graphs, enabling supervised training and systematic evaluation of models for HSQC peak prediction. What distinguishes 2DNMRGym is its dual-layer annotation strategy: the large-scale algorithm-generated annotations serve as silver-standard supervision for model training, while the expert-labeled subset provides a gold-standard benchmark to evaluate model robustness and generalization. This setup uses surrogate and abundant training labels to enable deep learning methods, and the high quality evaluation dataset to assess the ability of a model to learn meaningful molecular representations at the atom level. As such, the dataset offers a benchmark for existing and future GNN architectures in atom-level representation learning tasks.




\subsection{Concepts and terminology in chemistry}

\textbf{SMILES.}
Simplified Molecular Input Line Entry System (SMILES) \citep{weininger1988smiles} is a textual representation that employs short ASCII strings to describe chemical molecular structures. This notation system utilizes a series of characters, including alphanumeric symbols and punctuation marks, to represent the atoms, bonds, and connectivity within a molecule.

\textbf{Chemical shift.} Chemical shift is a measure of the resonant frequency of a nucleus relative to a reference standard, expressed in parts per million (ppm), and reflects the electronic environment surrounding the nucleus. In NMR spectroscopy, $^{1}$H chemical shifts typically range from 0 to 12 ppm, while $^{13}$C chemical shifts span a broader range, from 0 to 220 ppm, due to greater variation in carbon bonding environments. These shifts provide critical information about molecular structure, such as hybridization states, functional groups, and local electron density.

\textbf{HSQC.}
HSQC \citep{bodenhausen1980natural} is a 2D NMR spectroscopy technique used to elucidate the structure of molecules by correlating the chemical shifts of hydrogen atoms with those of directly bonded heteronuclei, typically carbon or nitrogen. This technique provides detailed insights into molecular connectivity and is particularly useful for studying complex organic compounds where traditional 1D NMR spectroscopy may not provide sufficient information. HSQC is instrumental in identifying atom-to-atom connections and understanding the molecular architecture of a substance.

\textbf{Tanimoto similarity.} Tanimoto similarity is a widely used metric in cheminformatics for comparing molecular fingerprints, which are typically represented as binary vectors \citep{bajusz2015tanimoto}. It quantifies the structural similarity between two molecules based on the presence or absence of shared substructures. 

\textbf{Scaffold.} Scaffold refers to the core structural backbone of a molecule, typically consisting of the ring systems and the connecting linkers, with side chains and substituents removed. It represents the central topology that defines a molecule’s overall shape and connectivity. In cheminformatics, scaffolds are often used to group molecules by structural similarity and to assess model generalization; for example, Bemis–Murcko scaffolds \citep{bemis1996properties} are commonly used to analyze scaffold diversity and enable tasks like scaffold splitting in molecular datasets.

\textbf{Hybridization.} Hybridization refers to the combination of atomic orbitals (e.g., \textit{sp}\textsuperscript{3}, \textit{sp}\textsuperscript{2}, \textit{sp}) to form new orbitals, which dictate the geometry of chemical bonds around an atom. This process affects both the electron distribution and the local chemical environment, factors that are crucial in determining NMR chemical shifts.

\textbf{Chirality.} Chirality is a molecular property where a compound exists as non-superimposable mirror images, usually due to a carbon atom bonded to four different substituents. This stereochemical feature affects the three-dimensional arrangement of atoms, which in turn influences the NMR signals, particularly in chiral environments.


\section{Related work}
The landscape of NMR databases exhibits a significant disparity in development and structure between 1D and 2D NMR spectra. For instance, the nmrshiftdb2 \citep{steinbeck2003nmrshiftdb} dataset provides a comprehensive collection of 1D data, serving as an open-access platform for the sharing of chemical shift information. This database is highly structured and extensively utilized across the computational chemistry community, making it a valuable resource for researchers. In contrast, databases that catalog 2D NMR spectra, such as those for HSQC, exhibit less cohesion and a greater degree of specialization, often tailored to specific sub-realms or applications within the field. The Human Metabolome Database (HMDB) \citep{wishart2022hmdb}, for example, is a rich resource that includes detailed HSQC spectra for thousands of metabolites, coupled with extensive metadata on their structures, biochemical properties, and roles in biological systems. This makes HMDB a vital tool for metabolomics research, aiding in the identification and detailed analysis of metabolites across various biological samples. Another dataset, CH-NMR-NP \citep{hayamizu2015open}, focuses on natural products and provides essential NMR spectral data, including HSQC spectra, for studying complex organic compounds. This dataset supports researchers in chemistry and biology by providing insights into the structure and potential applications of natural products, thus advancing the understanding of their biochemical pathways and therapeutic potentials. These specialized databases are not only repositories of NMR spectra but also rich sources of varied molecular dynamics and functional groups. Each database captures a unique slice of the chemical universe, encompassing a broad spectrum of molecular structures, which are represented as diverse graphs of varying sizes and complexities. This diversity is crucial for the development and evaluation of machine learning techniques, especially in the fields of computational chemistry and bioinformatics. While valuable, these databases were not designed with machine learning tasks in mind and lack the structured annotations necessary for supervised learning.

Furthermore, most existing ML models such as GCN~\citep{kipf2016semi}, GIN~\citep{xu2018powerful}, GAT~\citep{velickovic2017graph}, GNN Transformer~\citep{wu2021representing}, ComENet~\citep{wang2022comenet} and SchNet~\citep{schutt2018schnet} are trained at the molecule (graph-level) using coarse labels such as molecular properties using datasets like MolecularNet~\citep{wu2018moleculenet}, QMugs~\citep{isert2022qmugs}, GEOM~\citep{axelrod2022geom} etc., rather than capturing the finer atom-level interactions, as required in analyzing NMR spectra. Prior datasets rarely support this granularity, and those that do often rely on simulated data derived from quantum chemistry rather than real experimental spectra.

To address this gap, we introduce 2DNMRGym, a comprehensive, unified repository for experimental 2D NMR data. Unlike previous datasets, 2DNMRGym provides atom-level annotations, linking each cross peak to a specific hydrogen–heteronucleus bond within a molecular graph. The annotation process is labor-intensive and requires expert-level understanding of NMR and organic chemistry. To scale this effort, we adopt a dual-labeling strategy, combining algorithm-generated pseudo labels with a human-annotated subset for evaluation. This enables a unique atom-level representation learning task using surrogate supervision, where models are trained on imperfect algorithmic labels and evaluated against expert-labeled ground truth. In doing so, 2DNMRGym advances beyond traditional molecular fingerprinting and graph-level tasks, offering a new benchmark for fine-grained, chemically grounded prediction that bridges NMR spectroscopy and machine learning. This one-stop resource aims to streamline access and analysis of two-dimensional NMR spectra across various chemical contexts.


\section{Constructing the 2DNMRGym dataset}
\label{source}

Our 2DNMRGym dataset consists of over 22,000 HSQC spectra, where a small subset of 479 molecules with 7,310 cross peaks were randomly sampled for expert annotation as a held-out test set for evaluation. 

\begin{figure}[ht!]
    \centering
    \includegraphics[width=0.95\textwidth]{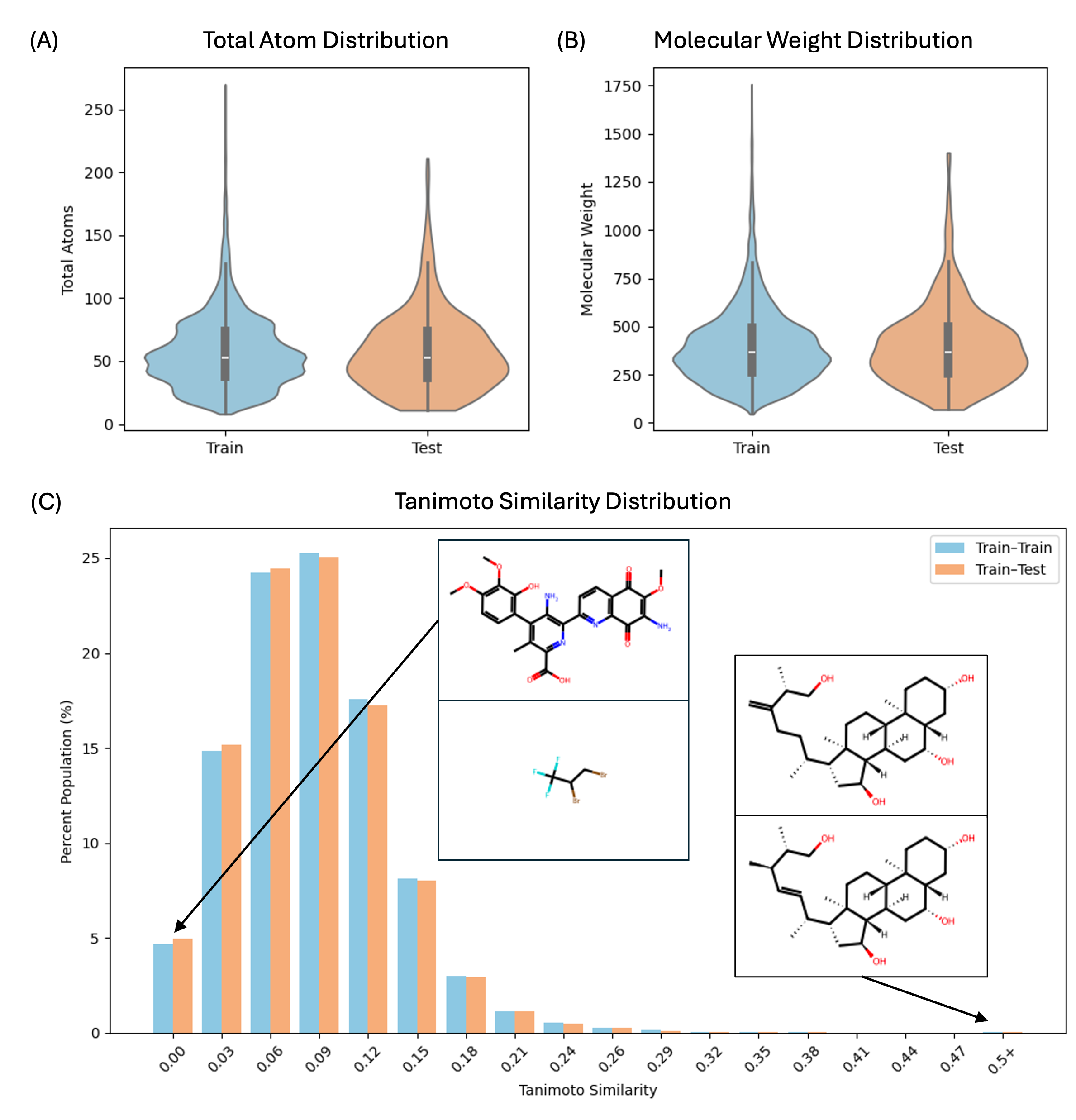} 
    \caption{Data statistics by number of atoms, molecular weight, and tanimoto similarity. }
    \label{fig:data_dist}
\end{figure}

Figure~\ref{fig:data_dist} summarizes key statistics of the training and test sets, which exhibit similar distributions in terms of total atom count, molecular weight, and Tanimoto similarity, indicating that the test set fairly represents the broader dataset and supports robust model evaluation. On average, molecules contain 58 atoms and have a molecular weight of approximately 400 Daltons. Over 25\% of the molecules exceed 75 atoms and 500 Daltons in weight. The Tanimoto similarity plot reveals that most molecule pairs have a similarity score below 0.1, highlighting the structural diversity of the dataset.

\begin{figure}[ht!]
    \centering
    \includegraphics[width=\textwidth]{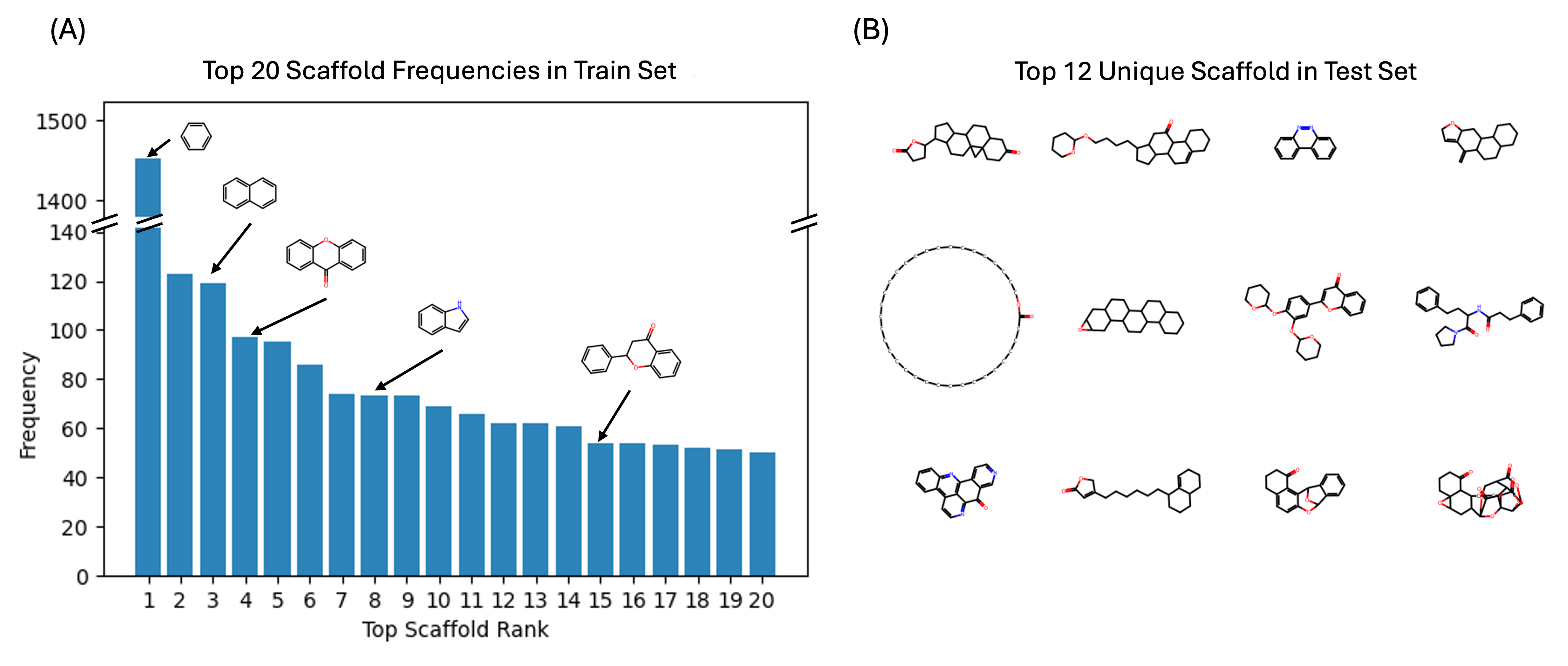} 
    \caption{Scaffold analysis for training and test dataset.}
    \label{fig:scaffold}
\end{figure}

To enable few-shot and zero-shot learning, we performed scaffold analysis for both the training and testing dataset. The test dataset contains 397 unique scaffolds, 148 of which are novel scaffolds that can be used for zero-shot learning. For scaffolds that appeared less than 10 times in the training set, they are used for few-shot learning. Figure \ref{fig:scaffold} summarizes the distribution and top scaffolds in the data.

\subsection{Collection of HSQC spectra and SMILES}
We meticulously curated 22,157 experimental spectra, along with NMR conditions and molecular CAS Registry numbers, which were extracted from the Human Metabolome Database(HMDB) \citep{wishart2022hmdb} (CC-NC-4.0 licence), and CH-NMR-NP \citep{hayamizu2015open} for each molecule were extracted from PubChem \citep{kim2023pubchem} (CC-BY-4.0 license) using their CAS Registry numbers.
The corresponding SMILES for each molecule were extracted from PubChem \citep{kim2023pubchem} using their CAS numbers. 

\subsection{Generation of molecular graphs}
Molecular graphs with stable 3D structures are derived from SMILES strings using the RDKit \citep{landrum2013rdkit} package, and formatted in Python Geometric format for computational processing. In the process of converting SMILES representations into molecular graphs, challenges arose with disjoint graphs, primarily due to the presence of floating ions. To ensure data quality and model accuracy, these anomalies are systematically identified and excluded from the dataset. Additionally, certain SMILES strings fail to yield energy-stable 3D structures despite multiple optimization attempts. These instances suggest structural inconsistencies or complexities that RDKit cannot resolve adequately. Such unstable entries are also eliminated to maintain the structural integrity and reliability of our dataset. This meticulous preprocessing ensures that our dataset only includes high-quality, consistent molecular graphs that are suitable for subsequent analysis and modeling.

Furthermore, using the RDKit \citep{landrum2013rdkit} package, we enrich the molecular graphs with node and edge features to infuse domain-specific insights into our Chemistry-Informed ML development. Three features are provided for each node: atomic type, chirality, and hybridization. Also, two features are considered for each edge: bond type and bond direction. Bond types include Single, Double, Triple, and Aromatic, each reflecting a distinct configuration of electron sharing between atoms. Bond direction includes None, EndUpRight, and EndDownRight, primarily representing stereochemistry in double bonds. ML practitioners have the option to incorporate these hand-crafted, domain-specific features in the model training process, which not only helps in understanding how traditional chemical knowledge translates into computational predictions but also explores how machine learning techniques can uncover patterns and relationships that might elude conventional domain expertise. This dual approach allows our models to benefit from established chemical theory while potentially discovering novel insights into molecular behavior that could redefine our understanding of NMR shifts and molecular interactions. Such findings could provide valuable contributions to the field, suggesting new areas of research or improvements to existing chemical theories.

\subsection{Annotation process}
\paragraph{Silver-standard labels} We use a framework proposed in \citep{li2025transpeaknet} to generate pseudo lables for 21,869 molecules. This model was first trained on extensive 1D NMR data, which establishes a robust foundation for understanding basic molecular interactions and chemical shift patterns. Afterwards, the model was fine-tuned on a diverse set of 2D NMR data, enhancing its ability to generalize across different molecular structures and solvent environments. With an accurate prediction of 2D NMR cross peaks, the model uses a matching algorithm to assign the predicted cross peaks to the most plausible observed peaks in the HSQC spectra, thus creating a direct linkage between each observed peak and its corresponding C--H bonds within the molecular graph. To test its annotation capability, we compared the annotation generated by this model to the expert annotations on our test dataset. Table \ref{annotation_accuracy} displays the result. Out of the 479 test molecules, the algorithm accurately annotates all peaks for 456 of the molecules (95.21\%). For the remaining 23 molecules, the model was able to annotate 81.56\% of the peaks accurately. 

\begin{table}[h]
\centering
\renewcommand{\arraystretch}{1.2}  
\caption{Pseudo-label Accuracy}
\label{annotation_accuracy}
\begin{tabular}{|c|c|}
\hline
\textbf{Fully-Correct Molecule (\%)} & \textbf{Peak Accuracy (\%) for Partial-Correct Molecule} \\
\hline
95.21\% & 81.56\% \\
\hline
\end{tabular}
\end{table}

\paragraph{Golden-standard labels}
The test dataset, comprising 479 molecules, underwent a rigorous multi-step annotation and validation process involving three domain experts to ensure the accuracy and reliability of labels used for model evaluation. The experts all have more than 10 years of experience in Organic Chemistry and NMR analysis, from Harvard University, Boston College and University of Georgia. Initially, all molecules were annotated by Expert A. Afterwards, the dataset was split into two subsets, each independently annotated and cross-checked by Expert B and Expert C. In cases of disagreement between the initial and secondary annotations, the molecule was flagged and reviewed by the third expert to resolve inconsistencies. The final consensus annotation agreed upon by at least two experts was recorded as the ground truth.

\section{2DNMRGym benchmark}
\begin{figure}[ht!]
    \centering
    \includegraphics[width=\textwidth]{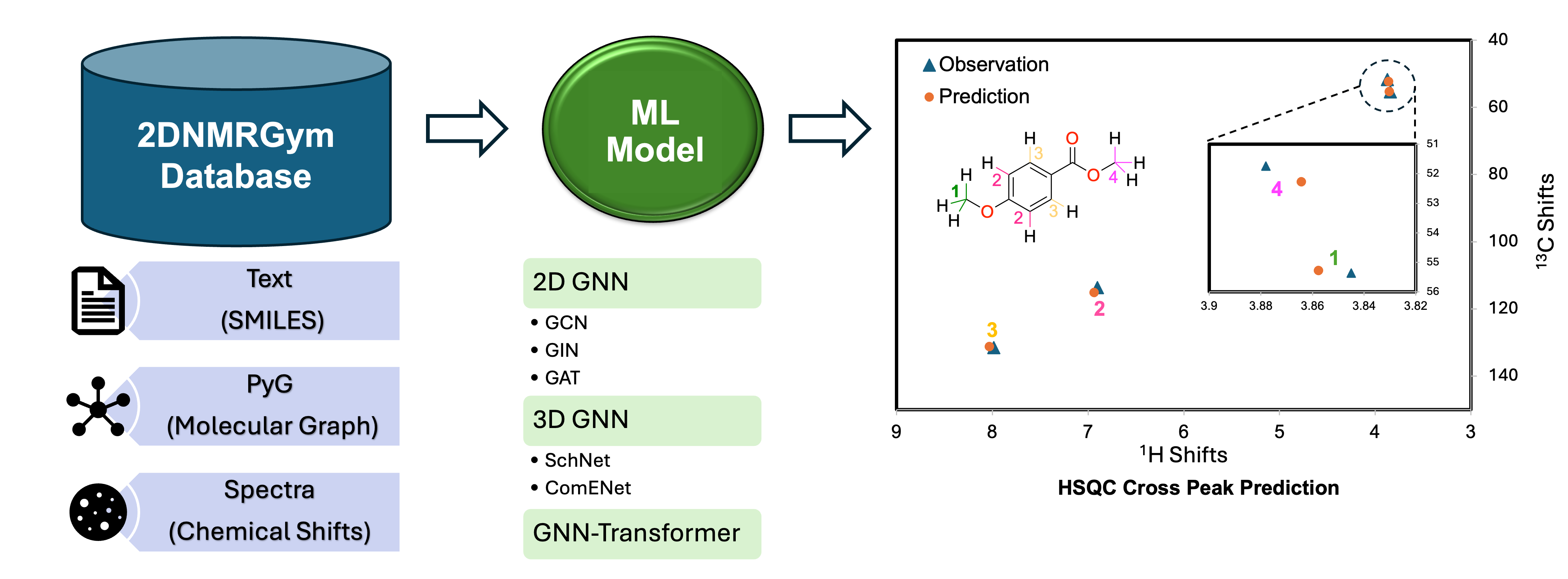}
    \caption{A demonstration workflow using 2DNMRGym dataset to train GNN models. The learnt graph representation from these benchmark models can be evaluated in the downstream HSQC cross peak prediction task.}
    \label{fig:task}
\end{figure}

To guide Machine Learning (ML) practitioners using 2DNMRGym, we provide benchmarks for cross peak prediction, an atom level representation learning task, described in Section \ref{intro} and Figure \ref{fig:task}. Models are evaluated on the held-out test set annotated by domain experts to ensure high-quality assessment. In addition to overall performance, we report results under few-shot and zero-shot evaluation settings to assess generalization. Specifically, a test molecule is considered few-shot if its scaffold appears fewer than 10 times in the training set, and zero-shot if its scaffold is not observed at all during training.

\subsection{Baseline models}
To benchmark atom-level cross-peak prediction, we evaluate several representative GNN architectures. For 2D GNNs, we include GCN \citep{kipf2016semi}, which performs neighborhood aggregation with normalized message passing; GIN \citep{xu2018powerful}, designed for maximal expressive power in distinguishing graph structures; and GAT \citep{velickovic2017graph}, which introduces attention mechanisms to weight neighbor contributions adaptively. We also incorporate GNN-Transformer \citep{wu2021representing}, a hybrid model combining GNNs with global self-attention and structural encodings to capture both local and long-range dependencies, which has shown strong performance on chemical and biological benchmarks. For 3D molecular graphs, we consider SchNet \citep{schutt2018schnet}, which leverages continuous-filter convolutions to model spatial interactions, and ComENet \citep{wang2022comenet}, which ensures full utilization of 3D geometric information within a 1-hop neighborhood. Together, these models provide a diverse baseline for evaluating atom-level representation learning on our 2DNMRGym dataset. The model details are included in Appendix C.

\subsection{Training and evaluation}
\paragraph{Train/validation split}
In our experiments, the data is randomly split into 80\% for training, 20\% for model selection, and the expert-annotated test dataset is used for model evaluation. For each model, we repeat the experiments using random seeds of 0, 42 and 66 and report the mean and standard deviatiion of Mean Absolute Error (MAE). 
\paragraph{Pre-processing}
The value ranges of the $^{13}\text{C}$- and $^{1}\text{H}$-shifts are quite different, 0 - 200 ppm for $^{13}\text{C}$ versus 0-12 ppm for $^{1}\text{H}$. To reduce bias and achieve better training, we normalized them to make their value range comparable by dividing $^{13}\text{C}$-shifts by 200 and $^{1}\text{H}$-shifts by 10. 
\paragraph{Error measurement}

As 2D NMR captures atomic interactions in two dimensions, specifically $^{13}$C-shift and $^{1}$H-shift, the model is trained using the Mean Absolute Error (MAE) of $^{13}$C-shifts and $^{1}$H-shifts, assigning them equal weights. The evaluation of the model's performance for both shifts is conducted using the MAE values calculated from the original values of the $^{13}$C- and $^{1}$H-shifts without normalization. This approach ensures that the model's predictions are assessed directly against the experimental chemical shift values, without any scaling or normalization, providing an unbiased assessment of its predictive capabilities for the two types of atomic interactions captured in 2D NMR spectra.



\subsection{Benchmark results}
All experiments were run using one V100 GPU. The performance of the baseline models is summarized in Table \ref{tab:gnn_comparison}. For each model, we adjusted its hyperparameters, including the hidden dimensions for GNN node representations, the hidden dimensions for edge representations (where applicable), the number of GNN layers, and the hidden channels of MLP layers for $^{13}\text{C}$-shifts and $^{1}\text{H}$-shifts predictions. Additionally for ComENet, we tune the number of layers inside the interaction module for node and edges during message passing. For SchNet, we also tune the number of filters in its filter-generating network. All models in this experiment are trained for 100 epochs with batch size set to 32.

\begin{table}[ht]
\centering
\begin{tabular}{p{1.8cm} p{1.7cm} p{1.2cm} p{1.2cm} p{1.2cm} p{1.2cm} p{1.2cm} p{1.2cm}}
\toprule
\textbf{Model Type} & \textbf{Model} & 
\multicolumn{2}{c}{\textbf{All-test MAE}} & 
\multicolumn{2}{c}{\textbf{Few-shot MAE}} & 
\multicolumn{2}{c}{\textbf{Zero-shot MAE}} \\
\cmidrule(lr){3-4} \cmidrule(lr){5-6} \cmidrule(lr){7-8}
& & \textbf{$^{13}\text{C}$} & \textbf{$^{1}\text{H}$}
  & \textbf{$^{13}\text{C}$} & \textbf{$^{1}\text{H}$}
  & \textbf{$^{13}\text{C}$} & \textbf{$^{1}\text{H}$} \\
\midrule
\multirow{3}{*}{2D GNN} 
& GCN       & 3.035 (0.039) & 0.229 (0.002) & 3.014 (0.011) & 0.227 (0.001) & 3.103 (0.038) & 0.242 (0.002) \\ 
& GIN       & 2.370 (0.007) & 0.203 (0.003) & \textbf{2.274 (0.022)} & 0.192 (0.002) & \textbf{2.587 (0.005)} & 0.230 (0.003) \\ 
& GAT       & 2.574 (0.045) & 0.206 (0.004) & 2.524 (0.042) & 0.201 (0.003) & 2.811 (0.066) & 0.226 (0.003) \\ 
\midrule
\multirow{2}{*}{3D GNN} 
& ComENet   & 3.143 (0.018) & 0.238	(0.003) & 3.178	(0.015) & 0.233	(0.002)& 3.348	(0.042)& 0.262	(0.003)\\ 
& SchNet    & 3.156 (0.022) & 0.240 (0.001) & 3.183 (0.014) & 0.239 (0.001) & 3.369 (0.031) & 0.261 (0.001) \\ 
\midrule
\multirow{3}{*}{Transformer} 
& GCN-Trans & 2.911 (0.044) & 0.221 (0.003) & 2.869 (0.036) & 0.215 (0.004) & 3.017 (0.055) & 0.241 (0.004) \\ 
& GIN-Trans & \textbf{2.348 (0.031)} & \textbf{0.198 (0.000)}& 2.281 (0.016) & \textbf{0.188 (0.001)} & 2.620 (0.039) & \textbf{0.228 (0.003)} \\ 
& GAT-Trans & 2.543 (0.097) & 0.206 (0.005) & 2.493 (0.104) & 0.200 (0.006) & 2.740 (0.079) & 0.228 (0.005) \\ 
\bottomrule
\end{tabular}
\caption{Comparison of MAE in ppm for $^{13}\text{C}$ and $^{1}\text{H}$ chemical shift predictions across different GNN models. The best model parameters are documented in Appendix D.}
\label{tab:gnn_comparison}
\end{table}

For all GNN models, adding the transformer component in model architecture generally boosts performance and reduces variances. Among GNN architectures, GIN models perform the best in our task due to their strong discriminative power, which is essential for capturing subtle structural variations that influence NMR shifts. Unlike GCN and GAT, GIN uses injective aggregation functions that better preserve node uniqueness within molecular graphs. Compared to GAT models, GIN is also architecturally simpler and tends to be more robust, especially when the dataset contains noise or biases introduced by silver standard labeling. This robustness makes GIN more reliable in learning meaningful representations from limited or noisy training data.

HSQC spectra primarily reflect short-range correlations governed by the 2D molecular structure, such as connectivity, atom types, hybridization, and chirality. These features, which are directly encoded in our graph representations, are sufficient to capture the stereoelectronic environments that determine chemical shifts. In contrast, 3D models like ComENet or SchNet rely on atomic coordinates that may not be optimal, as a molecule can adopt many possible conformers in solution. When only a single RDKit-embedded conformer is used, 3D models risk learning from spurious geometrical patterns or overfitting to noise in the 3D structure, leading to degraded performance compared to 2D models.

\section{Discussion and conclusion}
Our curated 2DNMRGym dataset is the first experimental, centralized, annotated, and high-quality dataset for learning atom-level molecular representation in the 2D NMR space. Significant effort was invested in the database's construction, with the cross-validation from three domain experts. Our dataset includes multimodal inputs such as text and graphs, and covers a wide range of molecules of varying sizes and scaffolds, providing valuable insights for evaluating representation learning models. To establish benchmark results, we tested a variety of 2D and 3D GNN models to predict HSQC cross peaks from molecular topologies/structures, paving the way for more advanced machine learning models for predicting HSQC cross peaks. The benchmarking results indicate that GIN stands out among the 2D and 3D GNN models that we have tried. This highlights the potential for developing 3D GNN models to capture spatial information such as chirality centers and hybridization, for atom-level tasks, which is potentially a major advance in NMR spectroscopy. There is plenty of room for improvements in prediction precision, aiming for an ideal MAE of less than 2 ppm for $^{13}\text{C}$ and less than 0.1 ppm for $^{1}\text{H}$.

Currently, the database contains only HSQC experimental data, which was generated to interrogate C--H interactions. Nevertheless, we expect the models trained on this HSQC data can be easily adapted or fine-tuned for other types of 2D NMR data. 
Looking ahead, the 2DNMRGym dataset is poised for further expansion to include a broader range of NMR techniques, such as HMBC and COSY, which probe different aspects of atomic interactions within molecules. Such expansions will enable the development of more advanced ML techniques for analyzing a wider array of NMR spectra, facilitating a more integrated approach to molecular characterization. 

\section{Acknowledgment}
This work was supported by GlycoMIP, a National Science Foundation (NSF) Materials Innovation Platform funded through Cooperative Agreement DMR-1933525, as well as NSF OAC 1920147. We also want to thank all the expert annotators: Dr. Hao Xu from Harvard Medical School, Dr. Duo-Sheng Wang from Boston College, and Dr. Ambrish Kumar from University of Georgia, Athens.

\newpage






\bibliography{neurips} 

\providecommand*{\mcitethebibliography}{\thebibliography}
\csname @ifundefined\endcsname{endmcitethebibliography}
{\let\endmcitethebibliography\endthebibliography}{}
\begin{mcitethebibliography}{33}
\providecommand*{\natexlab}[1]{#1}
\providecommand*{\mciteSetBstSublistMode}[1]{}
\providecommand*{\mciteSetBstMaxWidthForm}[2]{}
\providecommand*{\mciteBstWouldAddEndPuncttrue}
  {\def\EndOfBibitem{\unskip.}}
\providecommand*{\mciteBstWouldAddEndPunctfalse}
  {\let\EndOfBibitem\relax}
\providecommand*{\mciteSetBstMidEndSepPunct}[3]{}
\providecommand*{\mciteSetBstSublistLabelBeginEnd}[3]{}
\providecommand*{\EndOfBibitem}{}
\mciteSetBstSublistMode{f}
\mciteSetBstMaxWidthForm{subitem}
{(\emph{\alph{mcitesubitemcount}})}
\mciteSetBstSublistLabelBeginEnd{\mcitemaxwidthsubitemform\space}
{\relax}{\relax}

\bibitem[Gunther and Gunther(1994)]{gunther1994nmr}
H.~Gunther and H.~Gunther, \emph{NMR spectroscopy: basic principles, concepts, and applications in chemistry}, John Wiley \& Sons Chichester, UK, 1994\relax
\mciteBstWouldAddEndPuncttrue
\mciteSetBstMidEndSepPunct{\mcitedefaultmidpunct}
{\mcitedefaultendpunct}{\mcitedefaultseppunct}\relax
\EndOfBibitem
\bibitem[Claridge(2016)]{claridge2016high}
T.~D. Claridge, \emph{High-resolution NMR techniques in organic chemistry}, Elsevier, 2016, vol.~27\relax
\mciteBstWouldAddEndPuncttrue
\mciteSetBstMidEndSepPunct{\mcitedefaultmidpunct}
{\mcitedefaultendpunct}{\mcitedefaultseppunct}\relax
\EndOfBibitem
\bibitem[Yu \emph{et~al.}(2021)Yu, Myoung, and Ahn]{yu2021recent}
H.-Y. Yu, S.~Myoung and S.~Ahn, \emph{Magnetochemistry}, 2021, \textbf{7}, 121\relax
\mciteBstWouldAddEndPuncttrue
\mciteSetBstMidEndSepPunct{\mcitedefaultmidpunct}
{\mcitedefaultendpunct}{\mcitedefaultseppunct}\relax
\EndOfBibitem
\bibitem[Bodenhausen and Ruben(1980)]{bodenhausen1980natural}
G.~Bodenhausen and D.~J. Ruben, \emph{Chemical Physics Letters}, 1980, \textbf{69}, 185--189\relax
\mciteBstWouldAddEndPuncttrue
\mciteSetBstMidEndSepPunct{\mcitedefaultmidpunct}
{\mcitedefaultendpunct}{\mcitedefaultseppunct}\relax
\EndOfBibitem
\bibitem[Bross-Walch \emph{et~al.}(2005)Bross-Walch, K{\"u}hn, Moskau, and Zerbe]{bross2005strategies}
N.~Bross-Walch, T.~K{\"u}hn, D.~Moskau and O.~Zerbe, \emph{Chemistry \& biodiversity}, 2005, \textbf{2}, 147--177\relax
\mciteBstWouldAddEndPuncttrue
\mciteSetBstMidEndSepPunct{\mcitedefaultmidpunct}
{\mcitedefaultendpunct}{\mcitedefaultseppunct}\relax
\EndOfBibitem
\bibitem[Li and Kang(2020)]{li2020practical}
Q.~Li and C.~Kang, \emph{Molecules}, 2020, \textbf{25}, 2974\relax
\mciteBstWouldAddEndPuncttrue
\mciteSetBstMidEndSepPunct{\mcitedefaultmidpunct}
{\mcitedefaultendpunct}{\mcitedefaultseppunct}\relax
\EndOfBibitem
\bibitem[Kwon \emph{et~al.}(2020)Kwon, Lee, Choi, Kang, and Kang]{kwon2020neural}
Y.~Kwon, D.~Lee, Y.-S. Choi, M.~Kang and S.~Kang, \emph{Journal of chemical information and modeling}, 2020, \textbf{60}, 2024--2030\relax
\mciteBstWouldAddEndPuncttrue
\mciteSetBstMidEndSepPunct{\mcitedefaultmidpunct}
{\mcitedefaultendpunct}{\mcitedefaultseppunct}\relax
\EndOfBibitem
\bibitem[Yang \emph{et~al.}(2021)Yang, Chakraborty, and White]{yang2021predicting}
Z.~Yang, M.~Chakraborty and A.~D. White, \emph{Chemical science}, 2021, \textbf{12}, 10802--10809\relax
\mciteBstWouldAddEndPuncttrue
\mciteSetBstMidEndSepPunct{\mcitedefaultmidpunct}
{\mcitedefaultendpunct}{\mcitedefaultseppunct}\relax
\EndOfBibitem
\bibitem[Han \emph{et~al.}(2022)Han, Kang, Kang, Kwon, Lee, and Choi]{han2022scalable}
J.~Han, H.~Kang, S.~Kang, Y.~Kwon, D.~Lee and Y.-S. Choi, \emph{Physical Chemistry Chemical Physics}, 2022, \textbf{24}, 26870--26878\relax
\mciteBstWouldAddEndPuncttrue
\mciteSetBstMidEndSepPunct{\mcitedefaultmidpunct}
{\mcitedefaultendpunct}{\mcitedefaultseppunct}\relax
\EndOfBibitem
\bibitem[Chen \emph{et~al.}(2024)Chen, Liang, Tan, Wu, and Lu]{chen2024gt}
H.~Chen, T.~Liang, K.~Tan, A.~Wu and X.~Lu, \emph{Journal of Cheminformatics}, 2024, \textbf{16}, 132\relax
\mciteBstWouldAddEndPuncttrue
\mciteSetBstMidEndSepPunct{\mcitedefaultmidpunct}
{\mcitedefaultendpunct}{\mcitedefaultseppunct}\relax
\EndOfBibitem
\bibitem[Xu \emph{et~al.}(2023)Xu, Zhou, and Hong]{xu2023enhancing}
H.~Xu, Z.~Zhou and P.~Hong, \emph{arXiv preprint arXiv:2311.13817}, 2023\relax
\mciteBstWouldAddEndPuncttrue
\mciteSetBstMidEndSepPunct{\mcitedefaultmidpunct}
{\mcitedefaultendpunct}{\mcitedefaultseppunct}\relax
\EndOfBibitem
\bibitem[Li \emph{et~al.}(2025)Li, Xu, Kumar, Wang, Heiss, Azadi, and Hong]{li2025transpeaknet}
Y.~Li, H.~Xu, A.~Kumar, D.-S. Wang, C.~Heiss, P.~Azadi and P.~Hong, \emph{Communications chemistry}, 2025, \textbf{8}, 51\relax
\mciteBstWouldAddEndPuncttrue
\mciteSetBstMidEndSepPunct{\mcitedefaultmidpunct}
{\mcitedefaultendpunct}{\mcitedefaultseppunct}\relax
\EndOfBibitem
\bibitem[Weininger \emph{et~al.}(1988)Weininger, Weininger, and Weininger]{weininger1988smiles}
D.~Weininger, A.~Weininger and J.~Weininger, \emph{J Chem Inf Comput Sci}, 1988, \textbf{28}, 31--36\relax
\mciteBstWouldAddEndPuncttrue
\mciteSetBstMidEndSepPunct{\mcitedefaultmidpunct}
{\mcitedefaultendpunct}{\mcitedefaultseppunct}\relax
\EndOfBibitem
\bibitem[Bajusz \emph{et~al.}(2015)Bajusz, R{\'a}cz, and H{\'e}berger]{bajusz2015tanimoto}
D.~Bajusz, A.~R{\'a}cz and K.~H{\'e}berger, \emph{Journal of cheminformatics}, 2015, \textbf{7}, 1--13\relax
\mciteBstWouldAddEndPuncttrue
\mciteSetBstMidEndSepPunct{\mcitedefaultmidpunct}
{\mcitedefaultendpunct}{\mcitedefaultseppunct}\relax
\EndOfBibitem
\bibitem[Bemis and Murcko(1996)]{bemis1996properties}
G.~W. Bemis and M.~A. Murcko, \emph{Journal of medicinal chemistry}, 1996, \textbf{39}, 2887--2893\relax
\mciteBstWouldAddEndPuncttrue
\mciteSetBstMidEndSepPunct{\mcitedefaultmidpunct}
{\mcitedefaultendpunct}{\mcitedefaultseppunct}\relax
\EndOfBibitem
\bibitem[Steinbeck \emph{et~al.}(2003)Steinbeck, Krause, and Kuhn]{steinbeck2003nmrshiftdb}
C.~Steinbeck, S.~Krause and S.~Kuhn, \emph{Journal of chemical information and computer sciences}, 2003, \textbf{43}, 1733--1739\relax
\mciteBstWouldAddEndPuncttrue
\mciteSetBstMidEndSepPunct{\mcitedefaultmidpunct}
{\mcitedefaultendpunct}{\mcitedefaultseppunct}\relax
\EndOfBibitem
\bibitem[Wishart \emph{et~al.}(2022)Wishart, Guo, Oler, Wang, Anjum, Peters, Dizon, Sayeeda, Tian, Lee,\emph{et~al.}]{wishart2022hmdb}
D.~S. Wishart, A.~Guo, E.~Oler, F.~Wang, A.~Anjum, H.~Peters, R.~Dizon, Z.~Sayeeda, S.~Tian, B.~L. Lee \emph{et~al.}, \emph{Nucleic acids research}, 2022, \textbf{50}, D622--D631\relax
\mciteBstWouldAddEndPuncttrue
\mciteSetBstMidEndSepPunct{\mcitedefaultmidpunct}
{\mcitedefaultendpunct}{\mcitedefaultseppunct}\relax
\EndOfBibitem
\bibitem[Hayamizu \emph{et~al.}(2015)Hayamizu, Asakura, and Kurimoto]{hayamizu2015open}
K.~Hayamizu, K.~Asakura and T.~Kurimoto, 57th Experimental Nuclear Magnetic Resonance Conference, Pittsburgh, PA, 2015\relax
\mciteBstWouldAddEndPuncttrue
\mciteSetBstMidEndSepPunct{\mcitedefaultmidpunct}
{\mcitedefaultendpunct}{\mcitedefaultseppunct}\relax
\EndOfBibitem
\bibitem[Kipf and Welling(2016)]{kipf2016semi}
T.~N. Kipf and M.~Welling, \emph{arXiv preprint arXiv:1609.02907}, 2016\relax
\mciteBstWouldAddEndPuncttrue
\mciteSetBstMidEndSepPunct{\mcitedefaultmidpunct}
{\mcitedefaultendpunct}{\mcitedefaultseppunct}\relax
\EndOfBibitem
\bibitem[Xu \emph{et~al.}(2018)Xu, Hu, Leskovec, and Jegelka]{xu2018powerful}
K.~Xu, W.~Hu, J.~Leskovec and S.~Jegelka, \emph{arXiv preprint arXiv:1810.00826}, 2018\relax
\mciteBstWouldAddEndPuncttrue
\mciteSetBstMidEndSepPunct{\mcitedefaultmidpunct}
{\mcitedefaultendpunct}{\mcitedefaultseppunct}\relax
\EndOfBibitem
\bibitem[Velickovic \emph{et~al.}(2017)Velickovic, Cucurull, Casanova, Romero, Lio, Bengio,\emph{et~al.}]{velickovic2017graph}
P.~Velickovic, G.~Cucurull, A.~Casanova, A.~Romero, P.~Lio, Y.~Bengio \emph{et~al.}, \emph{stat}, 2017, \textbf{1050}, 10--48550\relax
\mciteBstWouldAddEndPuncttrue
\mciteSetBstMidEndSepPunct{\mcitedefaultmidpunct}
{\mcitedefaultendpunct}{\mcitedefaultseppunct}\relax
\EndOfBibitem
\bibitem[Wu \emph{et~al.}(2021)Wu, Jain, Wright, Mirhoseini, Gonzalez, and Stoica]{wu2021representing}
Z.~Wu, P.~Jain, M.~Wright, A.~Mirhoseini, J.~E. Gonzalez and I.~Stoica, \emph{Advances in neural information processing systems}, 2021, \textbf{34}, 13266--13279\relax
\mciteBstWouldAddEndPuncttrue
\mciteSetBstMidEndSepPunct{\mcitedefaultmidpunct}
{\mcitedefaultendpunct}{\mcitedefaultseppunct}\relax
\EndOfBibitem
\bibitem[Wang \emph{et~al.}(2022)Wang, Liu, Lin, Liu, and Ji]{wang2022comenet}
L.~Wang, Y.~Liu, Y.~Lin, H.~Liu and S.~Ji, \emph{Advances in Neural Information Processing Systems}, 2022, \textbf{35}, 650--664\relax
\mciteBstWouldAddEndPuncttrue
\mciteSetBstMidEndSepPunct{\mcitedefaultmidpunct}
{\mcitedefaultendpunct}{\mcitedefaultseppunct}\relax
\EndOfBibitem
\bibitem[Sch{\"u}tt \emph{et~al.}(2018)Sch{\"u}tt, Sauceda, Kindermans, Tkatchenko, and M{\"u}ller]{schutt2018schnet}
K.~T. Sch{\"u}tt, H.~E. Sauceda, P.-J. Kindermans, A.~Tkatchenko and K.-R. M{\"u}ller, \emph{The Journal of Chemical Physics}, 2018, \textbf{148}, \relax
\mciteBstWouldAddEndPuncttrue
\mciteSetBstMidEndSepPunct{\mcitedefaultmidpunct}
{\mcitedefaultendpunct}{\mcitedefaultseppunct}\relax
\EndOfBibitem
\bibitem[Wu \emph{et~al.}(2018)Wu, Ramsundar, Feinberg, Gomes, Geniesse, Pappu, Leswing, and Pande]{wu2018moleculenet}
Z.~Wu, B.~Ramsundar, E.~N. Feinberg, J.~Gomes, C.~Geniesse, A.~S. Pappu, K.~Leswing and V.~Pande, \emph{Chemical science}, 2018, \textbf{9}, 513--530\relax
\mciteBstWouldAddEndPuncttrue
\mciteSetBstMidEndSepPunct{\mcitedefaultmidpunct}
{\mcitedefaultendpunct}{\mcitedefaultseppunct}\relax
\EndOfBibitem
\bibitem[Isert \emph{et~al.}(2022)Isert, Atz, Jim{\'e}nez-Luna, and Schneider]{isert2022qmugs}
C.~Isert, K.~Atz, J.~Jim{\'e}nez-Luna and G.~Schneider, \emph{Scientific Data}, 2022, \textbf{9}, 273\relax
\mciteBstWouldAddEndPuncttrue
\mciteSetBstMidEndSepPunct{\mcitedefaultmidpunct}
{\mcitedefaultendpunct}{\mcitedefaultseppunct}\relax
\EndOfBibitem
\bibitem[Axelrod and Gomez-Bombarelli(2022)]{axelrod2022geom}
S.~Axelrod and R.~Gomez-Bombarelli, \emph{Scientific Data}, 2022, \textbf{9}, 185\relax
\mciteBstWouldAddEndPuncttrue
\mciteSetBstMidEndSepPunct{\mcitedefaultmidpunct}
{\mcitedefaultendpunct}{\mcitedefaultseppunct}\relax
\EndOfBibitem
\bibitem[Kim \emph{et~al.}(2023)Kim, Chen, Cheng, Gindulyte, He, He, Li, Shoemaker, Thiessen, Yu,\emph{et~al.}]{kim2023pubchem}
S.~Kim, J.~Chen, T.~Cheng, A.~Gindulyte, J.~He, S.~He, Q.~Li, B.~A. Shoemaker, P.~A. Thiessen, B.~Yu \emph{et~al.}, \emph{Nucleic acids research}, 2023, \textbf{51}, D1373--D1380\relax
\mciteBstWouldAddEndPuncttrue
\mciteSetBstMidEndSepPunct{\mcitedefaultmidpunct}
{\mcitedefaultendpunct}{\mcitedefaultseppunct}\relax
\EndOfBibitem
\bibitem[Landrum(2013)]{landrum2013rdkit}
G.~Landrum, \emph{Release}, 2013, \textbf{1}, 4\relax
\mciteBstWouldAddEndPuncttrue
\mciteSetBstMidEndSepPunct{\mcitedefaultmidpunct}
{\mcitedefaultendpunct}{\mcitedefaultseppunct}\relax
\EndOfBibitem
\bibitem[Bremser(1978)]{bremser1978hose}
W.~Bremser, \emph{Analytica Chimica Acta}, 1978, \textbf{103}, 355--365\relax
\mciteBstWouldAddEndPuncttrue
\mciteSetBstMidEndSepPunct{\mcitedefaultmidpunct}
{\mcitedefaultendpunct}{\mcitedefaultseppunct}\relax
\EndOfBibitem
\bibitem[Wiitala \emph{et~al.}(2006)Wiitala, Hoye, and Cramer]{wiitala2006hybrid}
K.~W. Wiitala, T.~R. Hoye and C.~J. Cramer, \emph{Journal of Chemical Theory and Computation}, 2006, \textbf{2}, 1085--1092\relax
\mciteBstWouldAddEndPuncttrue
\mciteSetBstMidEndSepPunct{\mcitedefaultmidpunct}
{\mcitedefaultendpunct}{\mcitedefaultseppunct}\relax
\EndOfBibitem
\bibitem[Mills(2006)]{mills2006chemdraw}
N.~Mills, \emph{ChemDraw Ultra 10.0 CambridgeSoft}, 2006\relax
\mciteBstWouldAddEndPuncttrue
\mciteSetBstMidEndSepPunct{\mcitedefaultmidpunct}
{\mcitedefaultendpunct}{\mcitedefaultseppunct}\relax
\EndOfBibitem
\bibitem[Willcott(2009)]{willcott2009mestre}
M.~R. Willcott, \emph{MestRe nova}, 2009\relax
\mciteBstWouldAddEndPuncttrue
\mciteSetBstMidEndSepPunct{\mcitedefaultmidpunct}
{\mcitedefaultendpunct}{\mcitedefaultseppunct}\relax
\EndOfBibitem
\end{mcitethebibliography}
\bibliographystyle{rsc}

\newpage
\appendix

\section{Annotation challenges}
2D NMR annotation, which involves associating the chemical shifts of each atom pair with the observed signals from experiments, is a highly challenging task. Using the HSQC spectrum as an example, the signals observed in the 2D spectrum correspond to the chemical shifts of hydrogen atoms directly bonded to heteronuclei, typically $^{13}\text{C}$ or $^{15}\text{N}$. Annotating these signals requires accurately mapping the observed cross-peaks to specific hydrogen-heteronucleus pairs within the molecule. However, this process is complicated by several factors, including spectral overlap, signal degeneracy, and sensitivity to experimental conditions.

\begin{figure}[ht!]
    \centering
    \includegraphics[width=\textwidth]{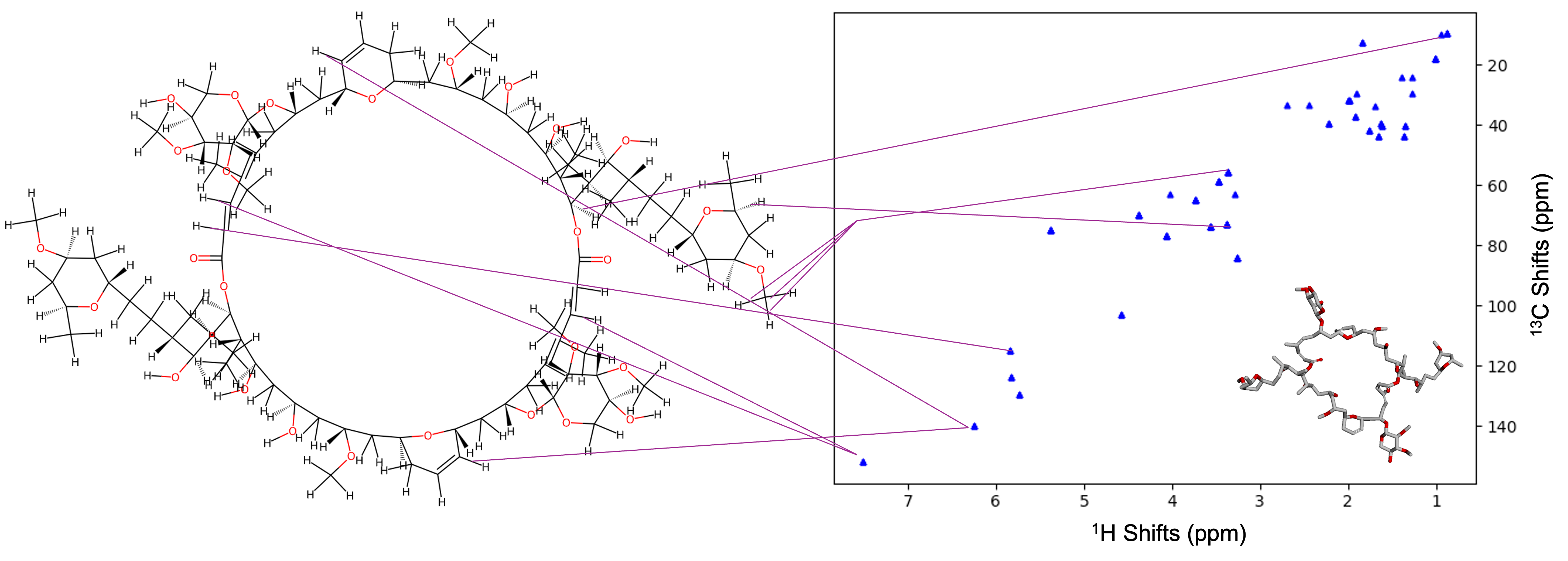}
    \caption{An annotation example. To avoid overcrowded, only a few ``C-H bond -- peak'' associations are shown. For a large molecule with complex structure like this, aligning the chemical bonds with the cross peaks is extremely difficult due to signal overlap and degeneracy. The bottom-right of the HSQC spectrum shows a 3D abstract skeleton of the molecule.}
    \label{fig:annotation}
\end{figure}

Spectral overlap occurs when multiple signals appear at similar chemical shift values, making it difficult to distinguish and assign them correctly. This issue is exacerbated in larger molecules with numerous hydrogen-heteronucleus pairs, leading to increased signal density and potential overlap. Additionally, signal degeneracy, where multiple atom pairs share the same chemical shift, further complicates the annotation process. Figure \ref{fig:annotation} shows an example of a large molecule in our dataset. 
Moreover, the observed chemical shifts are highly sensitive to the experimental conditions, such as temperature, solvent, pH, and sample concentration. Even slight variations in these conditions can cause detectable shifts in the signals, making it challenging to reliably match the experimental data with reference values or theoretical predictions. 

\section{Additional Concepts and terminology in chemistry}

\paragraph{Solvent}
A solvent, typically a liquid, is used to dissolve other substances (solutes), resulting in the formation of a solution. In the context of HSQC spectroscopy, solvent selection is paramount due to its profound influence on the chemical environment of the sample, thereby affecting the observed chemical shifts in NMR spectra. These shifts serve as pivotal indicators for accurately interpreting molecular structures as solvents can alter interactions such as hydrogen bonding, change molecular conformations, and affect the dynamics within a molecule. Thus, selecting an appropriate solvent and understanding its influence is essential for achieving precise and meaningful HSQC spectral analysis.

\paragraph{HOSE codes} 
HOSE \citep{bremser1978hose} codes are a method used in NMR spectroscopy for predicting chemical shifts. These codes function by encoding the structural environment of a nucleus in concentric spheres, capturing the types and positions of neighboring atoms up to several bonds away. Each sphere represents a distinct ``shell'' of neighbors, and the method relies on a database of known chemical shifts to predict the shift for a given atom based on its specific environment. This approach is empirical, utilizing accumulated historical data to make predictions.

\paragraph{DFT}
Density Functional Theory (DFT) \citep{wiitala2006hybrid} is a quantum mechanical method used to investigate the electronic properties of molecules and solids. In the context of NMR, DFT can be used to calculate chemical shifts by simulating the electronic environment around nuclei. This involves solving the Schrödinger equation for electrons in a molecule under the influence of a magnetic field, allowing for the prediction of NMR properties based on fundamental physical principles. DFT is known for its accuracy and ability to handle complex molecules, though it is computationally intensive compared to empirical methods like HOSE codes.
\paragraph{Traditional tools in  chemistry}
Two software tools are commonly used for processing, visualizing, simulating, and analyzing NMR spectral data, \textit{ChemDraw} \citep{mills2006chemdraw} and \textit{Mestrenova} \citep{willcott2009mestre}. They can serve as baselines for Machine Learning based methods.

\section{Benchmark GNN models}
\subsubsection{2D GNN models}

\paragraph{GCN} 
Graph Convolutional Networks (GCNs) \citep{kipf2016semi} is designed to efficiently learn node representations by leveraging the graph's structural information. The update rule for a GCN layer is formulated as follows:

\begin{equation}
    h^{(k+1)}_v = \sigma \left( W^{(k)} \sum_{u \in \mathcal{N}(v) \cup \{v\}} \frac{1}{\sqrt{\text{deg}(v) \text{deg}(u)}} h^{(k)}_u \right),
\end{equation}

where $h^{(k)}_v$ represents the feature vector of node $v$ at layer $k$, $\mathcal{N}(v)$ denotes the set of neighbors of node $v$, $W^{(k)}$ is the weight matrix at the $k$-th layer, and $\sigma$ is a non-linear activation function (e.g., ReLU), and $\text{deg}(v)$ and $\text{deg}(u)$ are the degrees of nodes $v$ and $u$, respectively. This approach, by normalizing based on node degrees, mitigates the problem of scale differences in node degrees, thus ensuring stable training and effective feature learning.

\paragraph{GIN}Graph Isomorphism Networks (GIN) \citep{xu2018powerful} are introduced to enhance the ability of GNNs to capture the structural nuances of graphs more effectively. Traditional GNN models often struggle to distinguish non-isomorphic graphs due to their limited expressiveness, akin to the Weisfeiler-Lehman (WL) graph isomorphism test. GINs are designed to address this issue by achieving maximal expressiveness in distinguishing graph structures. The general update rule for a GIN model is defined as follows:

\begin{equation}
    h^{(k+1)}_v = \text{MLP}^{(k)} \left( \left(1 + \epsilon^{(k)}\right) \cdot h^{(k)}_v + \sum_{u \in \mathcal{N}(v)} h^{(k)}_u \right),
\end{equation}

where $h^{(k)}_v$ is the feature vector of node $v$ at layer $k$, $\mathcal{N}(v)$ denotes the set of neighbors of node $v$, $\text{MLP}^{(k)}$ represents a multi-layer perceptron used at the $k$-th layer, $\epsilon^{(k)}$ is a learnable parameter or a fixed scalar that can be tuned to adjust the model's sensitivity to the central node's features.

\paragraph{GAT}
Graph Attention Networks (GATs) \citep{velickovic2017graph} incorporates the mechanism of attention into the GNN by dynamically assigning importance to nodes within a local neighborhood. The core update rule for a GAT model is expressed as follows:

\begin{equation}
    h^{(k+1)}_v = \sigma \left( \sum_{u \in \mathcal{N}(v) \cup \{v\}} \alpha_{vu}^{(k)} W^{(k)} h^{(k)}_u \right),
\end{equation}

where $h^{(k)}_v$ is the representation of node $v$ at layer $k$, $\mathcal{N}(v)$ denotes the neighbors of node $v$, $W^{(k)}$ is a weight matrix for the $k$-th layer, $\alpha_{vu}^{(k)}$ represents the attention coefficient between nodes $v$ and $u$, and $\sigma$ is a nonlinear activation function. The attention coefficients $\alpha_{vu}^{(k)}$ are computed through a learnable function of the features of nodes $v$ and $u$, allowing the model to focus more on relevant features during aggregation.

\paragraph{GNN transformer} The GNNTrans \citep{wu2021representing} model introduces a hybrid architecture that combines the expressive power of Graph Neural Networks (GNNs) with the global attention mechanism of Transformers to better capture both local and long-range dependencies in graph-structured data. By integrating structural encodings and a novel graph token, the model effectively handles graph-level tasks, achieving state-of-the-art performance on multiple benchmarks. This approach bridges the gap between sequential attention models and relational inductive biases in graphs. The model also achieves promising results on biological and chemical benchmarks, making it a suitable benchmark for our dataset. 

\subsubsection{3D GNN models}

\paragraph{ComENet}
ComENet \citep{wang2022comenet} offers an efficient message passing network designed specifically for 3D GNNs. It incorporates a new message passing scheme that ensures complete utilization of 3D information by operating within a 1-hop neighborhood, achieving both global and local completeness.

\paragraph{SchNet}
SchNet is another 3D GNN architecture designed for modeling atomic-scale interactions within molecules and materials \citep{schutt2018schnet}. It employs a unique continuous-filter convolutional approach to capture the complex interatomic forces and represents interatomic distances through a radial basis function expansion using a flexible number of Gaussian functions.

\section{Model parameters}
The optimal hyperparameters for each model in Table~\ref{tab:gnn_comparison} are summarized below. For each model type, extensive parameter tuning was conducted. The number of GNN layers tested included {3, 4, 5, 6}, with hidden dimensions of {256, 374, 512}. Prediction head configurations evaluated included {[256, 128], [128, 64], [256], [128]}. Solvent embedding dimensions were selected from {16, 32}. For the Transformer module, the hidden dimensions considered were {128, 256}, the number of attention heads {2, 3, 4}, feedforward dimensions {256, 512}, and the number of Transformer layers {3, 4, 5}.

\begin{table}[ht]
\centering
\caption{Model configurations for transformer GNN models}
\renewcommand{\arraystretch}{1.2}
\small
\begin{tabular}{p{0.5cm} p{0.5cm} p{0.5cm} p{0.5cm} p{1.2cm} p{1.2cm} p{1cm} p{1cm} p{0.5cm} p{0.5cm} p{0.5cm} p{0.5cm}}
\toprule
\textbf{Batch size} & \textbf{GNN type} & \textbf{GNN layer} & \textbf{Hid dim} & \textbf{Pred head (C)} & \textbf{Pred head (H)} & \textbf{Solvent emb (C)} & \textbf{Solvent emb (H)} & \textbf{Trans hid dim} & \textbf{Num of heads} & \textbf{Trans ff dim} & \textbf{Trans layer} \\
\midrule
32 & gin & 5 & 512 & {[128, 64]} & {[128, 64]} & 16 & 16 & 128 & 4 & 512 & 3 \\
32 & gcn & 5 & 512 & {[128, 64]} & {[128, 64]} & 16 & 16 & 128 & 4 & 256 & 5 \\
32 & gat & 5 & 512 & {[128, 64]} & {[128, 64]} & 16 & 16 & 128 & 2 & 512 & 5 \\
\bottomrule
\end{tabular}
\end{table}

\begin{table}[ht]
\centering
\caption{Model configurations for GNN-only models}
\renewcommand{\arraystretch}{1.2}
\small
\begin{tabular}{p{0.6cm} p{0.8cm} p{0.6cm} p{0.8cm} p{1.2cm} p{1.2cm} p{1cm} p{1cm} p{0.7cm} p{1.2cm}}
\toprule
\textbf{Batch size} & \textbf{GNN type} & \textbf{GNN layer} & \textbf{Hidden dim} & \textbf{Pred head (C)} & \textbf{Pred head (H)} & \textbf{Solvent emb (C)} & \textbf{Solvent emb (H)} & \textbf{Filters} & \textbf{Gaussians} \\
\midrule
32 & gat     & 5 & 512 & {[128, 64]} & {[128, 64]} & 32 & 16 & -- & -- \\
32 & gat     & 5 & 512 & {[128, 64]} & {[128, 64]} & 32 & 32 & -- & -- \\
32 & gcn     & 5 & 512 & {[128, 64]} & {[128, 64]} & 32 & 32 & -- & -- \\
32 & gin     & 5 & 512 & {[128, 64]} & {[128, 64]} & 32 & 16 & -- & -- \\
32 & gin     & 5 & 512 & {[128, 64]} & {[128, 64]} & 32 & 32 & -- & -- \\
32 & schnet  & 3 & 512 & {[128, 64]} & {[128, 64]} & 16 & 16 & 128 & 50 \\
32 & comenet & 6 & 512 & {[128, 64]} & {[128, 64]} & 16 & 16 & -- & -- \\
\bottomrule
\end{tabular}
\end{table}

\end{document}